\documentclass[11pt]{article}

\usepackage[letterpaper,margin=0.95in]{geometry}
\usepackage[T1]{fontenc}
\usepackage[utf8]{inputenc}
\usepackage{lmodern}
\usepackage{microtype}
\usepackage{float}
\usepackage{needspace}

\usepackage{amsmath,amssymb,amsthm,mathtools}
\usepackage{graphicx}
\usepackage{subcaption}
\usepackage{booktabs}
\usepackage{xcolor}
\usepackage[most]{tcolorbox}
\usepackage{hyperref}
\usepackage[numbers,sort&compress]{natbib}
\theoremstyle{definition}

\usepackage{algorithm}
\usepackage{algpseudocode}
\usepackage{multirow}
\usepackage{siunitx}

\definecolor{tildeBlack}{HTML}{0F0F0F}
\definecolor{tildeMuted}{HTML}{666666}
\definecolor{tildeLight}{HTML}{F2F3F5}
\definecolor{tildeBlue}{HTML}{1A73E8}

\hypersetup{
  colorlinks=true,
  linkcolor=black,
  citecolor=blue,
  urlcolor=black
}

\usepackage{pifont}

\usepackage{abstract} 
\newcommand{\tildelogo}{%
  \includegraphics[height=1.1em]{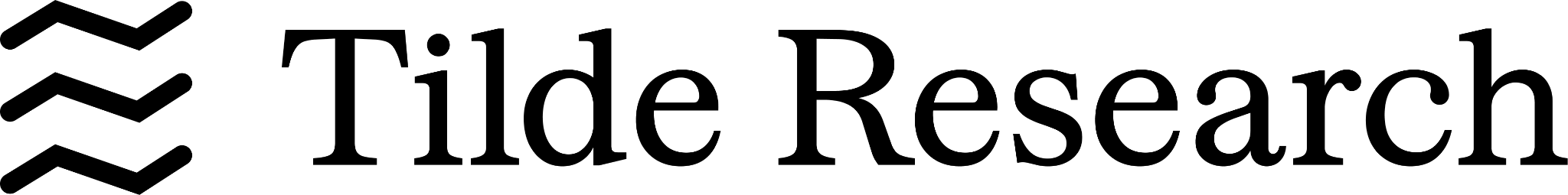}%
}
\newcommand{\northwesternlogo}{%
  \includegraphics[height=2.0em]{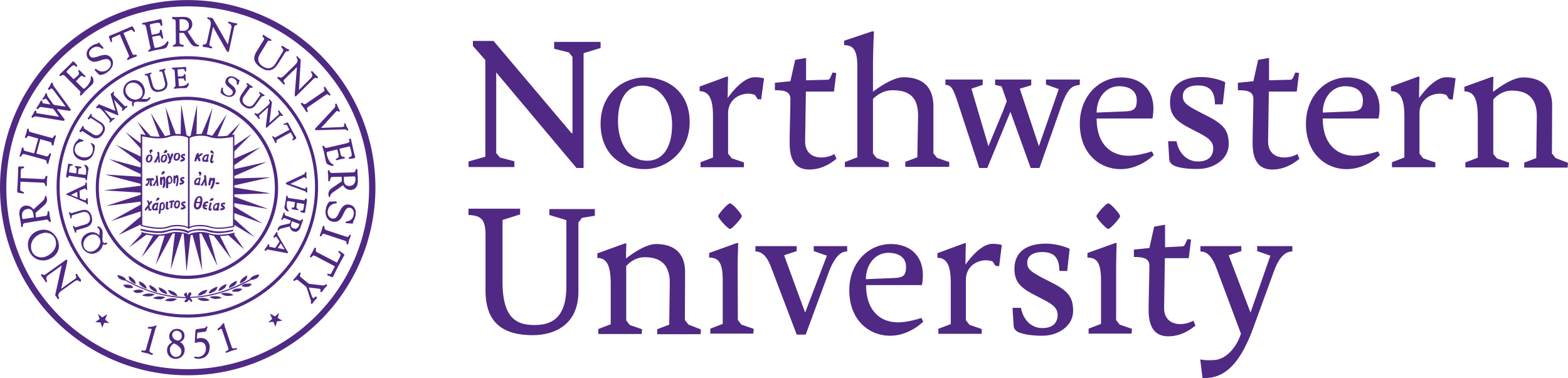}%
}

\newcommand{\papertitle}{Efficient and Trainable Language Model Test-Time Scaling via Local Branch Routing}
\newcommand{\paperdate}{June 2026}

\newcommand{\paperheader}{
\noindent
\begin{minipage}[c]{0.5\textwidth}
\northwesternlogo\hspace{1.2em}\tildelogo
\end{minipage}
\begin{minipage}[c]{0.5\textwidth}
\raggedleft{\large \paperdate}
\end{minipage}

\noindent\rule{\textwidth}{0.8pt}

\vspace{1.6em}

{\centering
\LARGE\bfseries
\papertitle\par
}

\vspace{0.8em}

\noindent\rule{\textwidth}{0.8pt}

\vspace{1.0em}

\begin{center}
{\large\bfseries
Yutong Yin$^{1}$,
Mingyu Jin$^{2}$,
Jin Pan$^{3,5}$,
Changyi Yang$^{4,5}$,
Zijie Xia$^{5}$,
Dhruv Pai$^{6}$,
Shuming Hu$^{6}$,
Zhen Zhang$^{7}$,
Chenyang Zhao$^{5}$,
Jinman Zhao$^{8}$,
Wujiang Xu$^{2}$,
Raymond Li$^{9}$,
Xin Eric Wang$^{7}$,
Julian McAuley$^{10}$,
Zhaoran Wang$^{1}$
\par}

\vspace{0.7em}

{\normalsize
$^{1}$Northwestern University,\;
$^{2}$Rutgers University,\;
$^{3}$University of Wisconsin--Madison,\;
$^{4}$Carnegie Mellon University,\;
$^{5}$LMSYS Org,\;
$^{6}$Tilde Research,\;
$^{7}$University of California, Santa Barbara,\;
$^{8}$University of Toronto,\;
$^{9}$University of British Columbia,\;
$^{10}$University of California, San Diego
\par}

\vspace{0.9em}

\begin{minipage}{0.86\textwidth}
\centering
Correspondence:
\href{mailto:yutongyin2028@u.northwestern.edu}{yutongyin2028@u.northwestern.edu},\\
\href{mailto:dhruv@tilderesearch.com}{dhruv@tilderesearch.com},\;
\href{mailto:zhaoranwang@gmail.com}{zhaoranwang@gmail.com}

\vspace{0.35em}
\textsc{Code:}
{\hypersetup{urlcolor=tildeBlue}%
\href{https://github.com/roger-yt/Local-Branch-Routing}
{\texttt{github.com/roger-yt/Local-Branch-Routing}}}
\end{minipage}
\end{center}

\vspace{0.6em}
}

\begin{document}

\paperheader

\begin{abstract}
Test-time scaling improves language-model reasoning, but existing approaches often face a difficult trade-off: long chain-of-thought sampling remains single-threaded, while sentence- or solution-level search can be computationally expensive and hard to train end-to-end. We introduce Local Branch Routing (LBR), a token-level test-time scaling framework that expands a small local lookahead tree, forwards all sampled branches through the language model, and uses a lightweight router to select the depth-1 subtree to commit. By routing over the hidden states of candidate local futures, LBR allows each token decision to use evidence beyond the root next-token distribution while avoiding full solution-level search. The resulting prune--shift--grow decoding process preserves discrete branch identities and defines a tractable tree-trajectory likelihood: newly grown nodes are counted when first sampled, and router decisions are assigned explicit probabilities. This enables end-to-end reinforcement learning with verifiable rewards, jointly optimizing the base model and router under the same likelihood-ratio principle as discrete-token RLVR. On synthetic hierarchical-planning tasks, LBR shows that post-candidate hidden states provide useful routing evidence. On mathematical reasoning benchmarks, LBR improves both Pass@1 and Pass@32 over discrete chain-of-thought, vanilla discrete-token RLVR, and RL-compatible soft-token branching baselines. These results suggest that lightweight local branching offers an efficient, trainable, and discrete form of language-model test-time scaling.
\end{abstract}

\vspace{2.0em}

% ------------------------------------------------------------
% Main paper
% ------------------------------------------------------------

\section{Introduction}

Recent progress in language-model reasoning shows that allocating more computation at inference time can substantially improve performance. Chain-of-thought prompting encourages models to externalize intermediate reasoning steps \citep{wei2022chain}, self-consistency improves accuracy by sampling multiple reasoning traces \citep{wang2022self}, and broader studies of test-time scaling show that the value of extra compute depends strongly on how it is allocated \citep{snell2024scaling}.
Search-based methods further introduce width by exploring trees of intermediate thoughts or actions \citep{yao2023tree, hao2023reasoning}, but such search is typically coarse-grained, computationally expensive, and difficult to optimize as a single trainable decoding policy.

This paper asks whether language models can obtain a lightweight form of test-time width at the token level. Standard autoregressive decoding commits to one next token before observing the hidden states induced by nearby alternatives. Yet in many reasoning problems, a local choice---such as a digit, operator, variable, or short phrase---can determine which latent state the model enters next. This suggests that candidate tokens should not only be scored by the root next-token distribution, but also by the local future states they induce.

We propose \emph{Local Branch Routing} (LBR), a trainable token-level branch-and-route decoding framework. At each decoding step, LBR expands a small local lookahead tree of width $K$ and depth $L$, forwards all sampled branches through the language model, and uses a lightweight router to select which depth-1 subtree to commit. The selected token is appended to the output, unselected subtrees are pruned, and the surviving subtree is shifted forward and regrown. Thus, each token decision can use post-candidate hidden states while avoiding full solution-level search.

LBR also defines a tractable tree-trajectory likelihood. The only stochastic operations are sampling newly grown tree nodes from the language model and sampling the router decision; pruning, shifting, and reuse are deterministic. This factorization allows LBR to be trained end-to-end with reinforcement learning from verifiable rewards, following the likelihood-ratio principle used in recent mathematical-reasoning systems \citep{shao2024deepseekmath, guo2025deepseek}.

Soft Thinking offers a related token-level branch-and-merge mechanism by replacing a committed
discrete token with a continuous mixture of candidate token embeddings~\citep{zhang2025soft}.
RL-compatible variants such as Multiplex Thinking add stochastic candidate sampling, making the
sampled candidates trainable with likelihood-ratio RL~\citep{tang2026multiplex}. However, these
methods merge candidates into a single soft token before future computation. LBR instead preserves
candidates as discrete forwarded branches, routes among their hidden states, and assigns explicit
probabilities to both tree growth and routing decisions.

\begin{figure}
  \centering
  \includegraphics[width=1.0\linewidth]{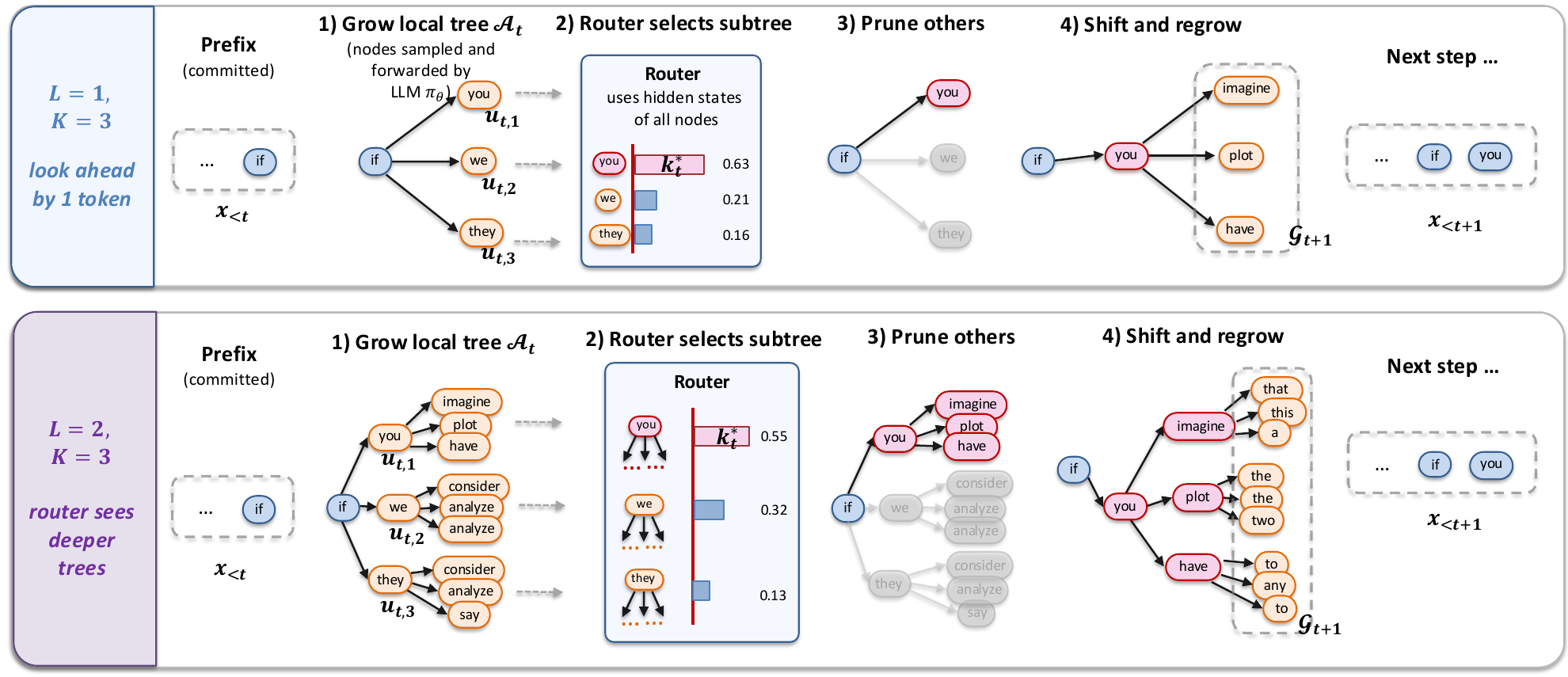}
\caption{
\textbf{Local Branch Routing decoding pipeline.}
LBR maintains a rolling local tree of already-forwarded candidate continuations. 
At each step, the router uses hidden states from all nodes in the current tree to select one depth-$1$ subtree, commits its root token, prunes the other subtrees, shifts the selected subtree forward, and grows one new layer to restore depth $L$. 
The top row shows the main experimental setting $L=1,K=3$, where routing uses post-token hidden states of $K$ candidate next tokens. 
The bottom row shows the same prune--shift--grow mechanism for $L=2,K=3$, where the router also observes hidden states of deeper local continuations.
}
\label{fig:lbr-decoding}
\end{figure}

We evaluate LBR on synthetic hierarchical planning and realistic mathematical reasoning. The synthetic task shows that post-candidate hidden states provide branch-specific information that is not available before branching. On math benchmarks, LBR improves over discrete chain-of-thought, vanilla RLVR, and RL-compatible soft-token branching baselines. These results suggest that local discrete branching offers an efficient and trainable form of language-model test-time scaling.

\section{Preliminaries}

Let $\mathcal V$ denote the vocabulary and let $\pi_\theta$ be an autoregressive language model.
Given a prefix $x_{<t}$, the model defines a next-token distribution
\(
    \pi_\theta(\cdot \mid x_{<t}) \in \Delta(\mathcal V).
\)
Let $E:\mathcal V\to\mathbb R^d$ be the token embedding map. After a token
$x_t\in\mathcal V$ is generated, its embedding $E(x_t)$ is fed back into the model for subsequent
computation.

We denote by
\(
    h_t = f_\theta(x_{<t}) \in \mathbb R^D
\)
the hidden state used to predict the next token at position $t$. The next-token logits are computed
from $h_t$, for example by $\mathrm{logits}_t = W_{\mathrm{out}}h_t$, and
\(
    \pi_\theta(\cdot \mid x_{<t})
    =
    \mathrm{softmax}(W_{\mathrm{out}}h_t).
\)
For tasks with verifiable answers, each prompt $q$ has a ground-truth answer $y^\star$ and a verifier
$v(y,y^\star)\in[0,1]$ that scores a generated response $y$. In math reasoning, this reward is often
binary, e.g.,
\(
    v(y,y^\star)=\mathbf 1\{\mathrm{Ans}(y)=y^\star\}.
\)
We use this verifiable reward later to train LBR with a likelihood-ratio objective over its tree
trajectory.

\section{Decoding Framework for Local Branch Routing}
\label{sec: decoding}

We introduce Local Branch Routing (LBR), a decoding framework that augments autoregressive
generation with a rolling local lookahead tree. As shown in Figure~\ref{fig:lbr-decoding}, each
decoding step follows four stages: grow a local tree, route among depth-1 subtrees, prune the
unselected subtrees, and shift-regrow the selected subtree. The tree is fully forwarded before routing, so each token decision uses hidden-state
evidence from local future continuations.

\paragraph{Stage 1: grow a forwarded local tree.}
Let $x_{<t}$ denote the committed prefix before decoding position $t$. Instead of sampling the next
token directly from $\pi_\theta(\cdot\mid x_{<t})$, LBR maintains a depth-$L$, width-$K$ local tree
$\mathcal{A}_t$ rooted at $x_{<t}$. Starting from the root, each active node samples $K$ children from
the filtered model distribution $\tilde{\pi}_\theta$, using the same temperature, top-$p$, top-$k$, or
other filters as ordinary decoding. For $L=1$, this gives $K$ candidate next tokens; for $L=2$, each
candidate next token is further expanded into $K$ one-step continuations.

Crucially, every sampled node is also forwarded through the language model. For each non-root node
$v\in\mathcal{A}_t$, let $\mathrm{path}(v)$ be the token sequence from the local-tree root to $v$. LBR
stores
\[
    h_t(v)=f_\theta(x_{<t},\mathrm{path}(v)).
\]
Thus $\mathcal{A}_t$ is not merely a set of candidate token strings; it is a forwarded tree of local
hidden states.
\begin{figure}[!t]
  \centering
  \includegraphics[width=1.0\linewidth]{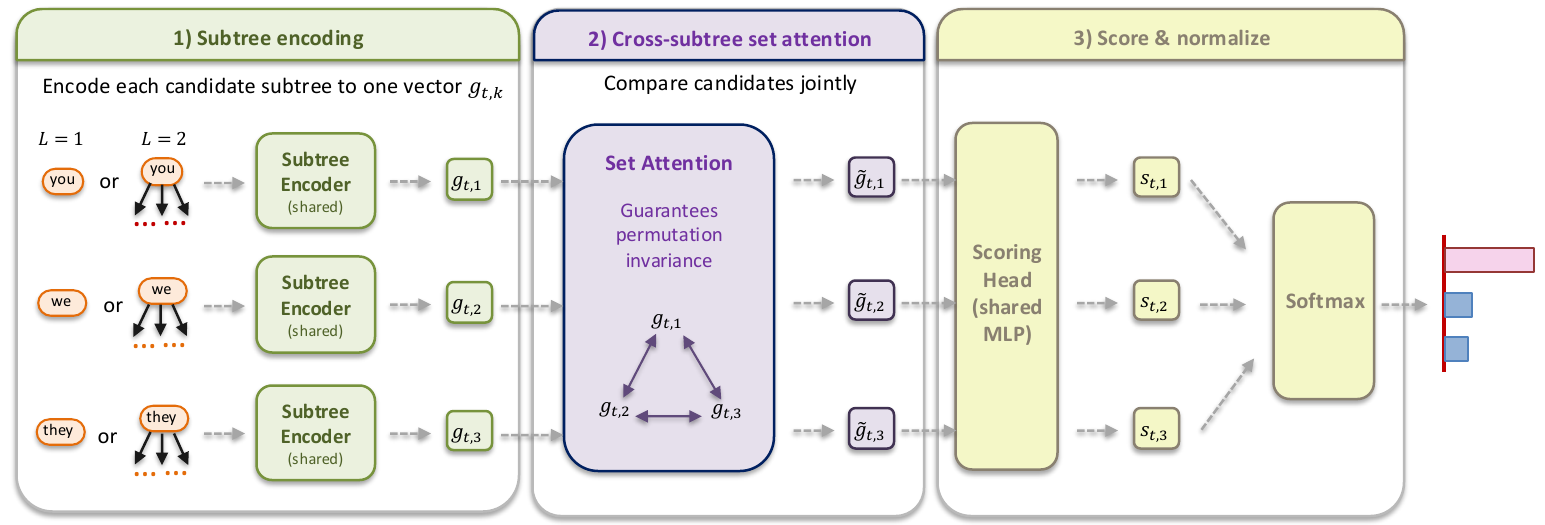}
\caption{
\textbf{Set-attention router.}
The router first encodes each depth-$1$ candidate subtree independently into a vector $g_{t,k}$.
For the main $L=1$ setting, a candidate subtree consists of a single forwarded token, so $g_{t,k}$ is computed from its post-token hidden state.
For $L=2$, the subtree encoder summarizes the hidden states of the candidate root and its local continuations.
The resulting candidate vectors are then passed through a cross-subtree set-attention block, where candidates attend to their sibling alternatives to produce context-aware representations $\tilde g_{t,k}$.
A shared scoring head maps these representations to scores $s_{t,k}$, and a softmax gives the routing distribution over candidate subtrees.
}
\label{fig:set-attention-router}
\end{figure}
\paragraph{Stage 2: route among depth-1 subtrees.}
Let $u_{t,1},\ldots,u_{t,K}$ be the depth-1 children of the root. Each $u_{t,k}$ defines one candidate
subtree, whose root token $\mathrm{tok}(u_{t,k})$ is a possible next committed token. The router
observes the forwarded tree and samples
\(
    k_t^\star \sim \rho_\phi(\cdot\mid x_{<t},\mathcal{A}_t;\theta),
\)
then commits
\(
    x_t=\mathrm{tok}(u_{t,k_t^\star}).
\)
The dependence on $\theta$ indicates that the router input contains hidden states produced by the
base language model.

We instantiate $\rho_\phi$ as the set-attention router in Figure~\ref{fig:set-attention-router}. Each depth-1
candidate subtree is first encoded into a vector $g_{t,k}$. When $L=1$, this is computed from the
post-token hidden state $h_t(u_{t,k})$; when $L>1$, the subtree encoder summarizes the candidate
root together with its local continuations. The candidate vectors are then compared by cross-subtree
set attention and scored by a shared head:
\[
    \tilde g_{t,1},\ldots,\tilde g_{t,K}
    =
    \mathrm{SetAttn}^{\mathrm{cand}}_\phi(g_{t,1},\ldots,g_{t,K}),
    \qquad
    s_{t,k}=w_\phi^\top \tilde g_{t,k},
\]
\[
    \rho_\phi(k\mid x_{<t},\mathcal{A}_t;\theta)
    =
    \frac{\exp(s_{t,k}/\tau)}
    {\sum_{k'=1}^{K}\exp(s_{t,k'}/\tau)}.
\]
This makes routing a relative decision over sibling local futures, rather than an independent
reranking of next-token logits. Additional architectural details are provided in Appendix \ref{subsec: router structure}.

\paragraph{Stage 3: prune unselected subtrees.}
After the router selects $k_t^\star$, LBR discards all depth-1 subtrees except the selected one. This
is the merge step: multiple local futures are explored, but only one subtree remains active. The
unselected branches no longer participate in future decoding, while the selected root token becomes
part of the committed prefix.

\paragraph{Stage 4: shift and regrow.}
The selected subtree is shifted forward: its root token has been committed, and its remaining
descendants become the partial lookahead tree for the next decoding position. LBR then grows one
new layer from the surviving frontier to restore depth $L$. 

We denote by $\mathcal{G}_t$ the set of
nodes newly grown and forwarded at step $t$. For $v\in\mathcal{G}_t$, let $\mathrm{tok}(v)$ be its
token label and define the tree-causal sampling context
\[
    \mathrm{ctx}(v)
    =
    \bigl(x_{<t},\mathrm{path}(\mathrm{pa}(v))\bigr),
\]
where $\mathrm{pa}(v)$ is the parent of $v$. Then
\(
    \mathrm{tok}(v)\sim \tilde{\pi}_\theta(\cdot\mid \mathrm{ctx}(v)).
\)

After an initial warm-up that builds the first depth-$L$ tree, decoding repeats the same rolling loop:
grow, route, prune, shift, and regrow.

\paragraph{Trace likelihood.}
This four-stage procedure defines a tractable likelihood for the full LBR decoding trace. At each
step, only two operations are stochastic: sampling newly grown nodes $\mathcal{G}_t$ and sampling
the router choice $k_t^\star$. The commit, prune, shift, and reuse operations are deterministic given
these samples. Let
\(
    \mathcal{F}
    =
    \{(\mathcal{G}_t,\mathcal{A}_t,k_t^\star)\}_{t=1}^{T}.
\)
Then
\begin{equation}
\label{eq:lbr-logprob}
    \log p_{\theta,\phi}(\mathcal{F}\mid q)
    =
    \sum_{t=1}^{T}
    \sum_{v\in\mathcal{G}_t}
    \log
    \tilde{\pi}_\theta
    \left(
        \mathrm{tok}(v)
        \mid
        \mathrm{ctx}(v)
    \right)
    +
    \sum_{t=1}^{T}
    \log
    \rho_\phi
    \left(
        k_t^\star
        \mid x_{<t},\mathcal{A}_t;\theta
    \right).
\end{equation}
The first term accounts for growing the local trees, and the second accounts for routing decisions.
The sum is over $\mathcal{G}_t$, not all of $\mathcal{A}_t$: a reused node may influence later
routers through its hidden state, but its language-model log-probability is counted only when it is
first grown.

\section{Interpreting Local Branch Routing on Synthetic Hierarchical Planning}

We hypothesize that LBR improves reasoning by exploiting predictive hidden states at branching
tokens to perform local planning. To illustrate this mechanism, we study a synthetic hierarchical
planning task based on radix-translated graph reachability. The experiment shows that LBR has an
advantage over discrete CoT and soft thinking because it effectively utilizes the informative post-branching hidden states by routing. Our synthetic task is modified from the ProsQA benchmark introduced by \citet{hao2024training} and further analyzed by \citet{zhu2025reasoning}.

\begin{figure}[!t]
  \centering
  \includegraphics[width=1.0\linewidth]{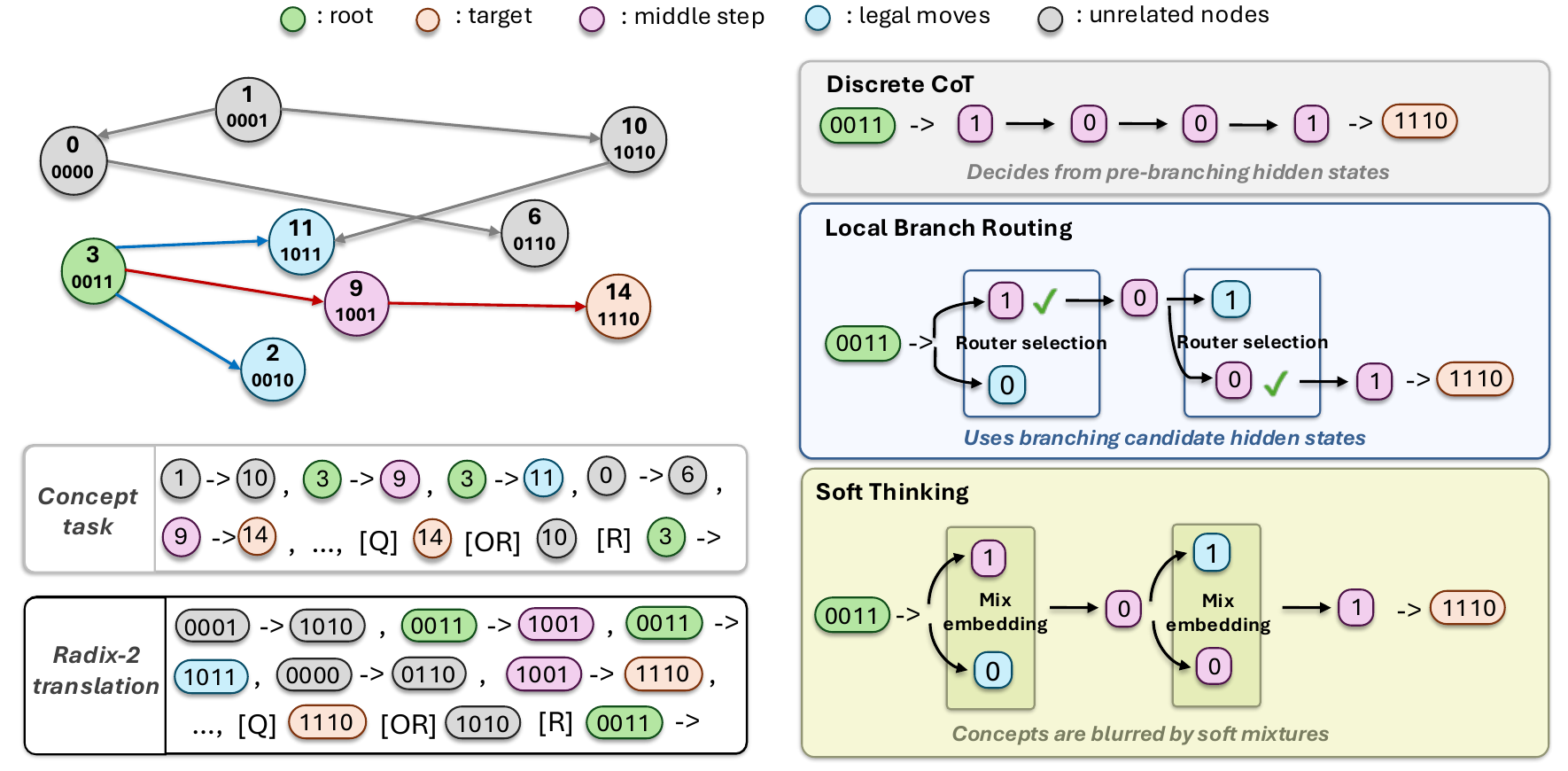}
\caption{
Radix-translated graph reachability and decoding behavior. Left: a concept-level
reachability problem is translated into a radix-2 token sequence, so each graph node becomes a
short digit string and each graph transition becomes a sequence of token-level decisions. Shared
digit prefixes create hierarchical branching positions. Right: the three decoding methods handle
these branching positions differently. Discrete CoT commits from the pre-branching hidden state,
LBR forwards candidate digits as separate branches and routes using their post-candidate hidden
states, and Soft Thinking mixes candidate embeddings into a single continuous state, which can
blur concept identity.
}
\label{fig:interp}
\end{figure}

\subsection{Radix-Translated Reachability as Hierarchical Planning}

The left panel of Figure~\ref{fig:interp} illustrates the synthetic planning task. At the
concept level, each example is a directed graph reachability problem: the model is given a serialized
graph, two candidate targets, and a root node, and must generate a path from the root to the reachable
target. In the example, the root is \(3\), the candidate targets are \(14\) and \(10\), and only \(14\) is
reachable. The legal next concepts from \(3\) are \(11,9,2\); a local search procedure must identify that
the branch through \(9\) reaches the target and generate the path \(3\to 9\to 14\).

We translate this concept-level problem into a radix-tokenized sequence by representing each node id
as a fixed-length binary string. For \(n=16\) nodes, each concept uses \(W=4\) digit tokens, e.g.,
\(3=0011\), \(9=1001\), \(11=1011\), and \(2=0010\). Thus a concept-level transition such as
\(3\to 9\) becomes a sequence of token-level decisions \(0011\to 1001\).

This translation creates hierarchical branching at the token level. The competing next concepts
\(11,9,2\) share and diverge across digit prefixes: the first digit separates the \(0***\) group from the
\(1***\) group; after choosing prefix \(1\), the next digit is temporarily forced to \(0\); a later digit then
distinguishes \(1001\) from \(1011\). Therefore, concept-level planning is converted into a sequence
of local token decisions with both merges and branch points, making the task a controlled testbed for
whether post-candidate hidden states help route among locally ambiguous branches.

\subsection{Training and Decoding Methods}

The right panel of Figure~\ref{fig:interp} compares three decoding methods on the same
radix-translated transition. \textbf{Discrete CoT} treats the path as a standard autoregressive
sequence and commits to each digit from the pre-branching hidden state. \textbf{Local Branch
Routing} keeps the same discrete token space, but forwards each legal candidate digit as a separate
branch and uses a router to select among the resulting post-candidate hidden states. \textbf{Soft
Thinking} exposes the same legal candidates, but merges them into a continuous mixture embedding
before future computation, thereby blurring the identity of the selected concept branch.

All methods use the same base model and data distribution. We first apply MTP pretraining to teach
the model legal graph transitions by predicting future node digits along random walks. We then
post-train each method with its corresponding trace representation: Discrete CoT uses the gold radix
path with next-token cross-entropy; LBR forwards the legal candidate branches and supervises the
router to select the gold branch; Soft Thinking trains on branch-and-merge traces where legal
candidate embeddings are mixed before predicting future tokens. Thus, the comparison controls for
graph knowledge and isolates how each method represents branching decisions.

\begin{figure}[t]
    \centering

    \begin{subfigure}[t]{0.49\linewidth}
        \centering
        \includegraphics[width=\linewidth]{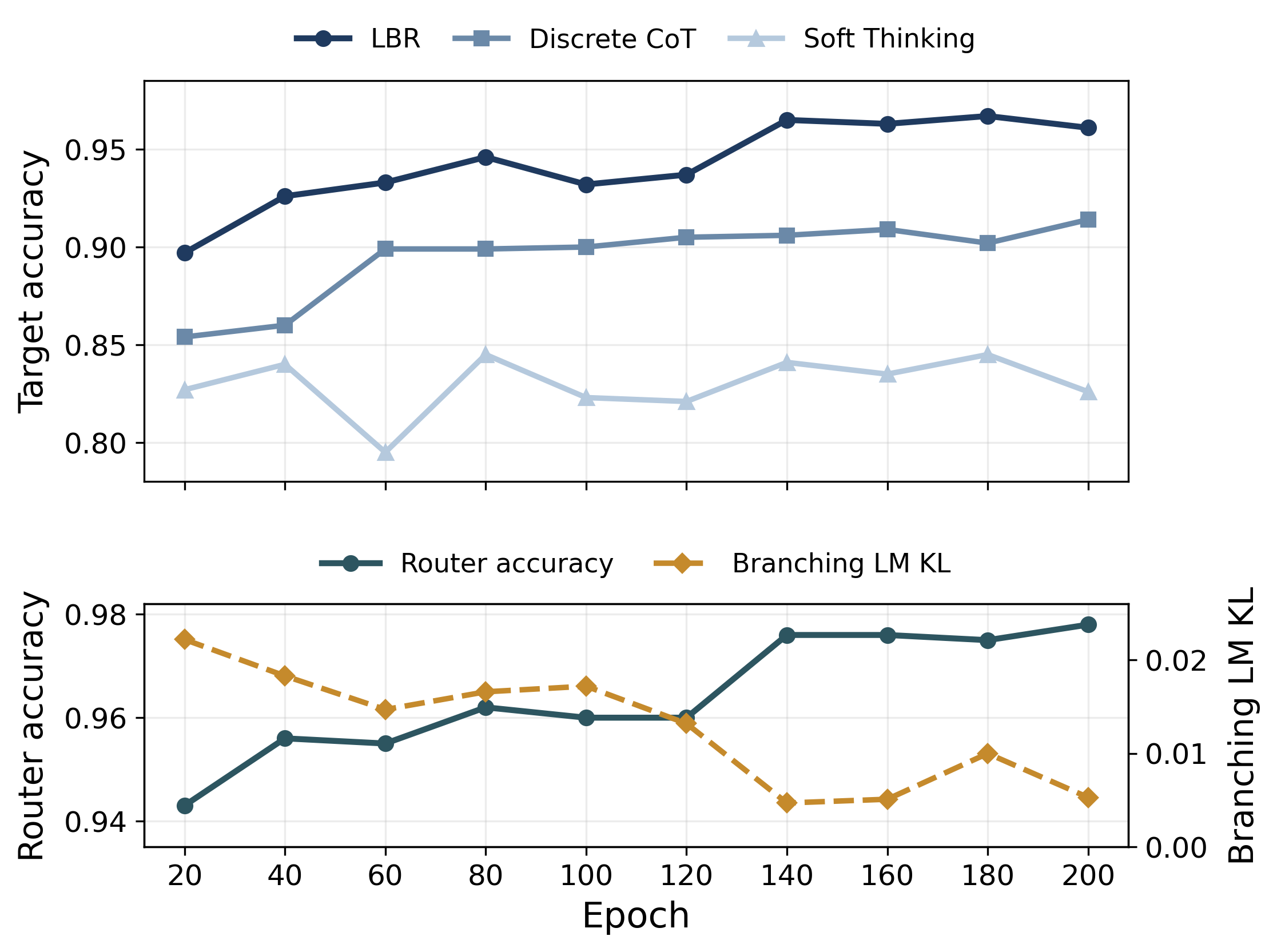}
        \caption{Learning dynamics.}
        \label{fig:learning-dynamics}
    \end{subfigure}
    \hfill
    \begin{subfigure}[t]{0.49\linewidth}
        \centering
        \includegraphics[width=\linewidth]{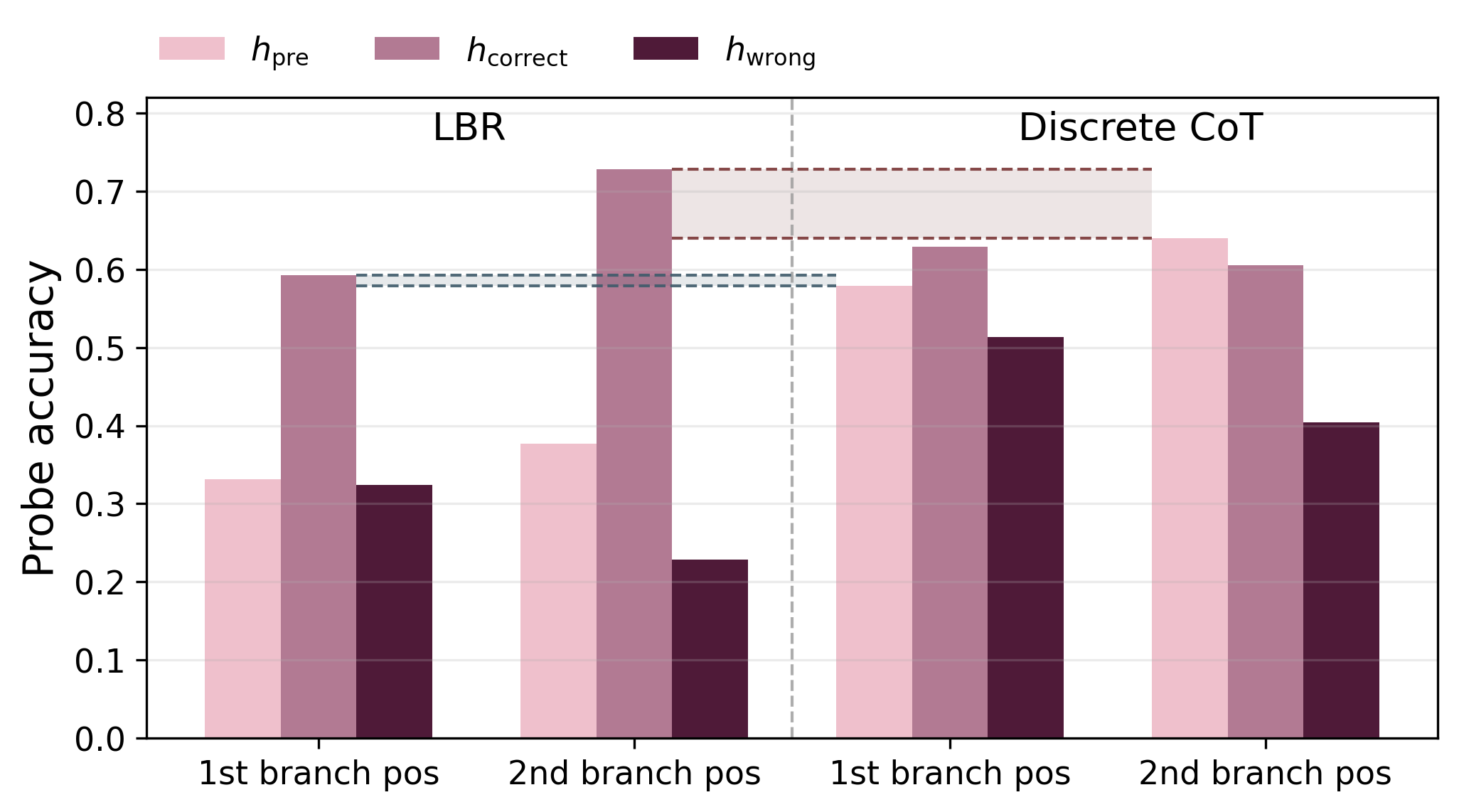}
        \caption{Branching-state probe.}
        \label{fig:branching-probe}
    \end{subfigure}

    \caption{
    Synthetic hierarchical-planning results.
    Left: LBR achieves the highest target accuracy on radix-translated reachability, while its
    router accuracy increases and branching-position LM KL decreases during training.
    Right: probes show that LBR's post-correct-candidate hidden states are more predictive of the
    final target than pre-branching states, supporting routing after forwarding candidate branches.
    }
    \label{fig:synthetic-learning-probe}
\end{figure}

\subsection{Why LBR Helps: Informative Branching Hidden States}

The upper panel of Figure~\ref{fig:synthetic-learning-probe}(a) compares the target accuracy of the three
methods. LBR achieves the best performance, followed by discrete CoT and then
Soft Thinking:
\[
    \text{Soft Thinking }(0.845)
    <
    \text{Discrete CoT }(0.914)
    <
    \text{LBR }(0.967).
\]
We next ask why routing after branching gives this advantage.

The key difference is the hidden state from which a branching decision is made. Discrete CoT must
choose the next digit from the pre-branching hidden state, before any candidate digit is realized. LBR
instead first forwards each legal candidate digit and then routes using the resulting post-candidate
hidden states. Figure~\ref{fig:synthetic-learning-probe}(b) tests whether these post-candidate states contain
more information about the final target. The probe predicts the target node id from hidden states at
diverging positions. For LBR, the post-correct-candidate state is much more predictive than the
pre-branching state. Thus, the candidate token itself reveals downstream information that is not
available before branching.

This also explains the gap to discrete CoT. Although discrete CoT learns to encode some decision information in its pre-branching state, this state
remains weaker than LBR's post-correct-candidate state, as highlighted by the shaded band in
Figure~\ref{fig:synthetic-learning-probe}(b). Moreover, there is a large separation between correct and
wrong post-candidate states. Therefore, the router is not
merely reranking equivalent candidates; it can exploit candidate-specific hidden states that distinguish
branches leading toward the target from branches leading away from it.

The lower panel of Figure~\ref{fig:synthetic-learning-probe}(a) confirms that LBR learns to use this signal
during training. Router accuracy on oracle branching events increases, while the branching-position
LM KL decreases. This indicates that the router becomes better at selecting the correct forwarded
branch, and the base model simultaneously learns a more calibrated local branching distribution.
Appendix~\ref{apd: soft ambig} provides an additional concept-identity probe showing that
soft-token mixtures blur the identity of the generated graph node, whereas LBR preserves discrete
candidate identities.

\section{Local Branch Routing Improves Mathematical Reasoning}
\paragraph{Experimental setup.}
We evaluate LBR on DeepSeek-R1-Distill-Qwen-1.5B and DeepSeek-R1-Distill-Qwen-7B. All
RL-trained methods use GRPO with verifier rewards. For LBR, we use the same RLVR setup as in
Section~3, replacing the standard discrete-token log-probability with the tree-trace likelihood in
Eq.~\ref{eq:lbr-logprob}:
\(
    \mathcal{L}_{\mathrm{LBR}}(\theta,\phi;\mathcal{F})
    =
    -A(\mathcal{F})\log p_{\theta,\phi}(\mathcal{F}\mid q).
\)

This assigns reward-weighted credit to both the newly grown local tree tokens and the router
decisions. When $K=1$, the router is deterministic and the objective reduces to standard
discrete-token RLVR.

\paragraph{Implementation.}
We jointly train the base model and router. The base-model learning rate is $1\times 10^{-6}$ and
the router learning rate is $1\times 10^{-4}$. The router samples candidate subtrees with temperature
$1.0$. We train for 300 steps with a global batch size of 128 questions and 8 rollout samples per
question. Training uses temperature $1.0$ and top-$p=1.0$; evaluation uses top-$p=0.95$ and reports
pass@1/32 averaged over 64 sampled runs. All training and evaluation use a maximum response length
of 4096 tokens. The training set is DeepScaleR-Preview-Dataset \citep{deepscaler2025}. All the trainings are done in 8$\times$H100 GPUs. Router architecture hyperparameters are provided in Appendix~\ref{apd: router hyper}.

\paragraph{Baselines.}
We compare with discrete chain-of-thought decoding, vanilla GRPO/RLVR with single-thread
discrete tokens, and RLVR-compatible soft-token branching methods such as Multiplex Thinking.
We report the main $L=1,K=3$ configuration and ablate deeper local lookahead such as $L=2,K=3$.

For vanilla discrete-token RLVR, a sampled response
$y=(y_1,\ldots,y_T)$ is trained with the standard likelihood-ratio loss
\(
    \mathcal L_{\mathrm{disc}}(\theta)
    =
    -A(q,y)
    \sum_{t=1}^T
    \log \pi_\theta(y_t\mid q,y_{<t}),
\)
where $A(q,y)$ is the advantage computed from verifier rewards.

For RL-compatible soft-token branching, following Multiplex Thinking, each step samples
$K$ candidate tokens $c_{t,1},\ldots,c_{t,K}$ and merges their embeddings into a soft token
$\tilde e_t$. Its sampled candidates define the training likelihood
\(
    \mathcal L_{\mathrm{soft}}(\theta)
    =
    -A(q,\tilde y)
    \sum_{t=1}^{T}
    \sum_{k=1}^{K}
    \log \pi_\theta(c_{t,k}\mid q,\tilde y_{<t}).
\)

\subsection{Math Benchmark Results}

Table~\ref{tab:math pass} reports Pass@1/Pass@32 on six mathematical reasoning benchmarks.
Across both 1.5B and 7B backbones, LBR improves over discrete CoT, vanilla discrete-token RLVR,
and RLVR with soft-token branching. This comparison isolates the decoding framework: the
RL-trained methods use the same verifier rewards, but LBR replaces single-thread or soft-merged
decoding with routing over forwarded local branches. The gains over soft-token branching show that
preserving candidate branches and routing among their hidden states is more effective than merging
candidate embeddings into a single continuous token.

LBR improves both Pass@1 and Pass@32, suggesting better single-sample accuracy as well as better
exploration under repeated sampling. Increasing the local lookahead from \(L=1\) to \(L=2\) often
further improves performance, especially on the 7B backbone, consistent with the view that deeper
local continuations provide more useful routing evidence.
\begin{table*}[t]
\centering
\caption{
Main mathematical reasoning results. We report Pass@1/Pass@32 on six benchmarks for
DeepSeek-R1-Distill-Qwen-1.5B and 7B. LBR improves over discrete CoT, vanilla discrete-token
RLVR, and RLVR-compatible soft-token branching baselines.
}
\label{tab:math pass}
\small
\renewcommand{\arraystretch}{1.10}
\setlength{\tabcolsep}{9pt}

\makebox[\textwidth][c]{%
\begin{tabular}{lcccccc}
\toprule
\textbf{Method} & \textbf{Minerva} & \textbf{AIME'25} & \textbf{AIME'24} & \textbf{MATH500} & \textbf{AMC'23} & \textbf{Olympiad} \\
\midrule

\multicolumn{7}{c}{\textbf{DeepSeek-R1-Distill-1.5B}} \\
\specialrule{0.08em}{0.10em}{0.25em}
Discrete CoT        & 23.4/61.8 & 11.3/32.0 & 10.2/32.9 & 66.3/91.6 & 36.9/74.8 & 30.5/60.5 \\
RLVR w. Discrete Tokens         & 24.3/62.4 & 12.0/32.1 & 10.5/31.8 & 66.7/91.1 & 38.4/74.6 & 31.2 /61.3\\
RLVR w. Soft Tokens   & 26.2/\textbf{63.6} & 12.8/34.7 & 11.8/40.3 & 67.5/\textbf{94.6} & 38.7/80.1 & 31.3/67.7\\
LBR ($L=1$)  & \textbf{29.7}/\textbf{64.2} & \textbf{15.8}/\textbf{37.7}& \textbf{15.1}/40.7 & \textbf{75.0}/93.7 & \textbf{48.1}/\textbf{86.0} & \textbf{38.2}/67.4 \\
LBR ($L=2$)  & \textbf{30.1}/\textbf{63.5} & \textbf{16.2}/34.9 & \textbf{14.8}/\textbf{42.9} & \textbf{75.5}/\textbf{94.5} & \textbf{49.2}/\textbf{84.8} & \textbf{39.0}/\textbf{68.3} \\
% inrouter ($d=1,w=3$)  & 28.8/62.2 & 14.6/37.2 & 14.3/41.7 & 73.9/93.4 & 46.3/82.6 & 37.6/66.7 \\
\specialrule{0.08em}{0.25em}{0.35em}

\multicolumn{7}{c}{\textbf{DeepSeek-R1-Distill-7B}} \\
\specialrule{0.08em}{0.10em}{0.25em}
Discrete CoT        & 33.3/65.5 & 16.0/32.7 & 15.7/40.8 & 71.6/93.7 & 42.4/78.5 & 35.6 /65.5\\
RLVR w. Discrete Tokens       & 35.3/65.7 & 17.1/31.3 & 17.2/43.5 & 74.1/93.7 & 44.7/80.7 & 38.0 /66.0\\
RLVR w. Soft Tokens   & 38.6/66.3 & 19.7/\textbf{40.5} & 20.6/57.9 & 78.0/95.4 & 50.7/87.4 & 41.7/70.2 \\
LBR ($L=1$)  & \textbf{41.9}/66.0 & 23.4/\textbf{40.8} & 23.9/55.0 & \textbf{83.1}/\textbf{96.8} & 57.3/\textbf{89.6} & 46.7/71.9 \\
LBR ($L=2$)  & \textbf{42.3}/\textbf{67.3} & \textbf{28.0}/\textbf{40.1} & \textbf{28.0}/\textbf{62.8} & \textbf{85.6}/\textbf{97.1} & \textbf{61.3}/88.7 & \textbf{50.2}/\textbf{74.4} \\
% inrouter ($d=1,w=3$)  & 39.2/65.7 & 21.8/43.4 & 21.3/52.4 & 79.6/95.1 & 52.9/87.4 & 43.8/70.2 \\
\specialrule{0.08em}{0.25em}{0.35em}
\bottomrule
\end{tabular}%
}
\end{table*}

\subsection{Cross-Subtree Attention Improves Routing}

\begin{figure}[t]
    \centering

    \begin{minipage}[t]{0.47\textwidth}
        \centering
        \includegraphics[width=\textwidth]{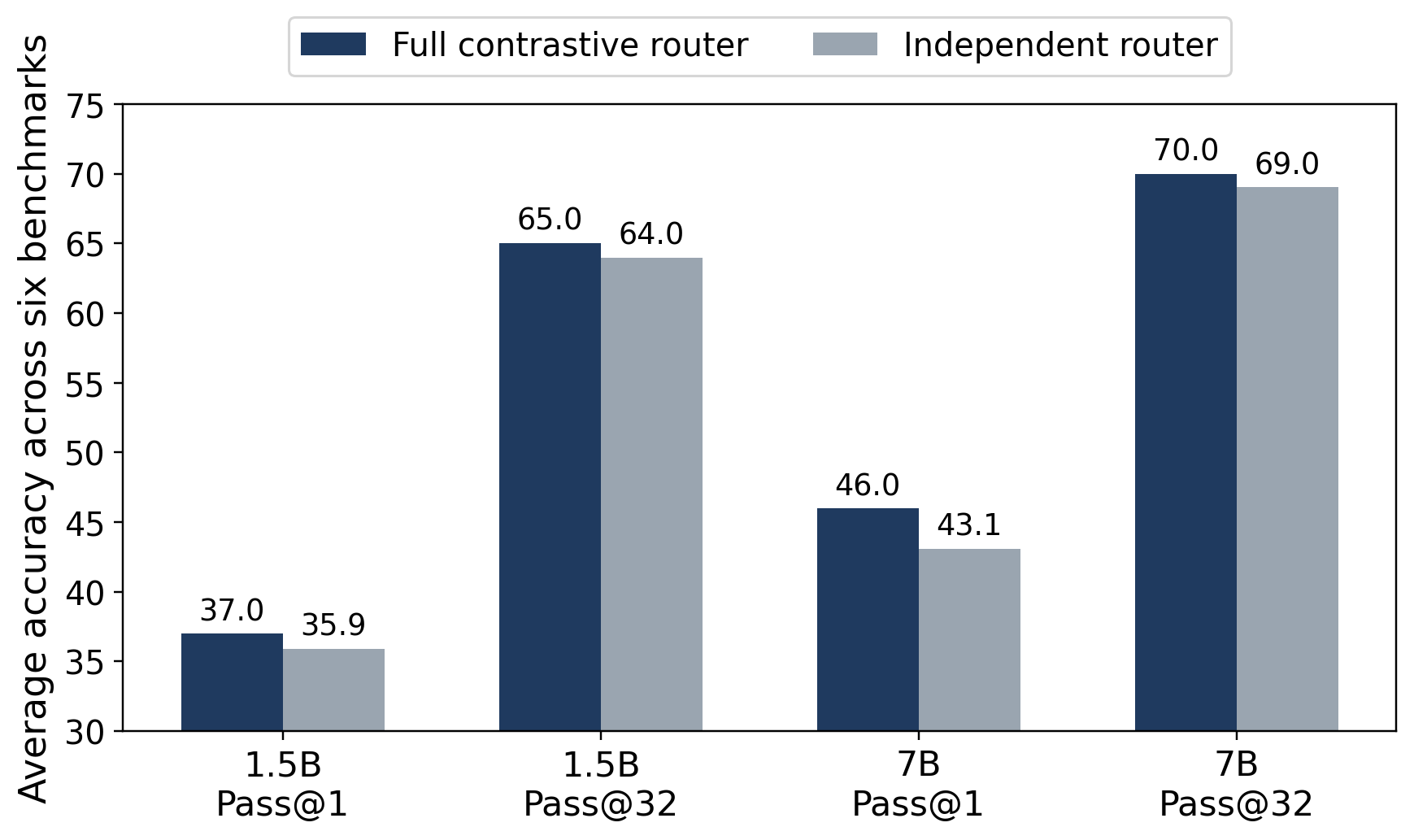}
        \vspace{-0.8em}
        \caption{
        Router ablation. The full contrastive router outperforms the independent router,
        showing that comparing sibling branches improves routing.
        }
        \label{fig:router-ablation}
    \end{minipage}
    \hfill
    \begin{minipage}[t]{0.47\textwidth}
        \centering
        \includegraphics[width=\textwidth]{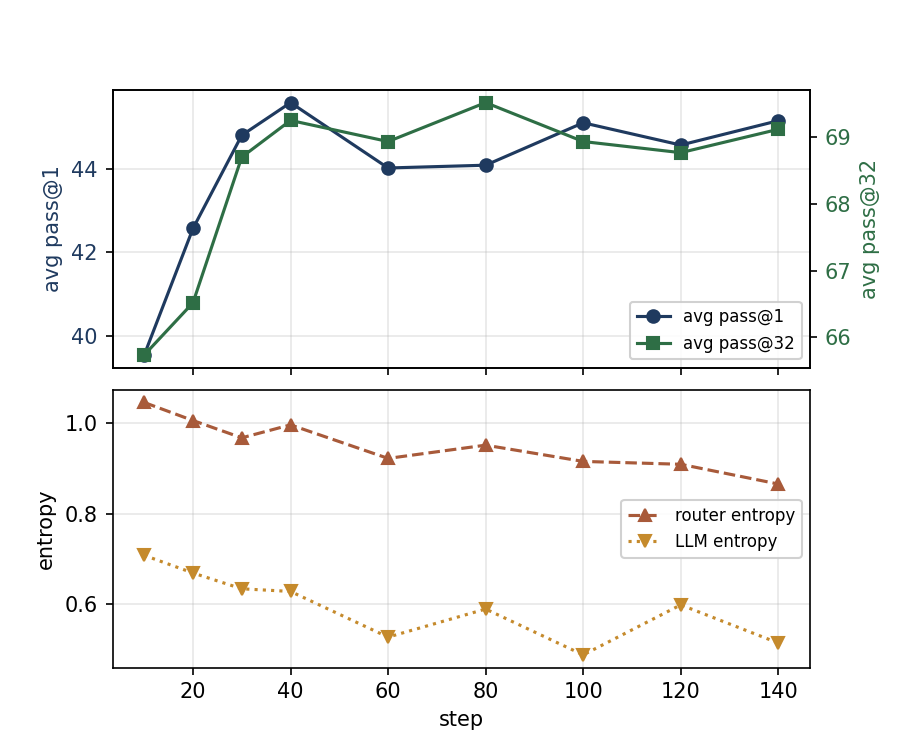}
        \vspace{-0.8em}
        \caption{
        Training dynamics of LBR during RLVR. Average Pass@1 and Pass@32 improve early and
        remain stable, while router entropy and base-LM entropy at routed positions decrease.
        }
        \label{fig:lbr-training-dynamics}
    \end{minipage}

    \vspace{-0.8em}
\end{figure}

Figure~\ref{fig:router-ablation}
compares the full set-attention router with an independent router. The independent variant removes
cross-subtree attention and scores each candidate subtree with a shared MLP applied to its own
embedding only. This keeps the same LBR decoding framework, tree width, training setup, and
tree-trace objective, but removes the contrastive comparison among sibling candidates.

The independent router consistently underperforms the full router. This gap shows that LBR's gains
are not explained only by exposing the router to post-token hidden states; the way those hidden states
are compared also matters. In particular, cross-subtree attention lets each candidate be evaluated
relative to the other sampled local futures from the same prefix. The ablation therefore supports the
claim that routing is a contrastive decision over local branches, rather than independent scoring of
candidate tokens.

\subsection{Training Dynamics of Local Branch Routing}

Figure~\ref{fig:lbr-training-dynamics} shows the training dynamics of LBR during RLVR on the
math-reasoning benchmarks.

The top panel reports average Pass@1 and Pass@32 across the evaluated tasks. The main improvement
happens early in training: average Pass@1 rises from roughly \(39.5\) to above \(45\), while average
Pass@32 rises from roughly \(65.5\) to around \(69\). After this early phase, both metrics remain
relatively stable, suggesting that LBR quickly learns a useful local routing policy and then maintains
its gains without severe degradation.

The entropy curves provide a complementary view of what is learned. Router entropy decreases
throughout training, indicating that the router becomes more decisive in selecting among local
branches. At the same time, the base-LM entropy at routed positions also decreases, suggesting that
the language model itself becomes more confident about local branch tokens. Thus, LBR training
does not only train an external selector on top of a fixed distribution; it co-adapts the base model and
router so that local candidate branches become easier to distinguish and route.

Importantly, router entropy decreases together with improved Pass@1 and Pass@32, rather than
through a collapse that hurts exploration. This supports the interpretation that LBR learns a
discriminative routing policy over forwarded local branches. The result also helps explain why LBR improves both single-sample
accuracy and Pass@32: the router becomes sharper, while the underlying sampling process still
maintains enough diversity across repeated runs.

\section{Related Works}

\paragraph{Test-time scaling and search.}
Test-time scaling improves language-model reasoning by allocating additional inference-time compute.
Chain-of-thought prompting and self-consistency scale reasoning through longer traces or multiple
sampled traces, while recent work studies budget forcing, verifier-guided search, and adaptive sampling
\citep{wei2022chain, wang2022self, snell2024scaling, muennighoff2025s1}. Search-based methods
introduce explicit width by exploring trees over thoughts, actions, or solution refinements, including
Tree of Thoughts, RAP, LATS, policy-guided tree search, and adaptive-branching MCTS
\citep{yao2023tree, hao2023reasoning, zhou2023language, li2025policy, inoue2025wider}.
These methods demonstrate the value of branching during inference, but typically operate over coarse
reasoning units and often rely on external evaluators, handcrafted search control, or repeated
full-step generation. LBR instead branches over token-level futures, forwards the resulting hidden
states, and routes to one depth-1 subtree before continuing autoregressive decoding.

\paragraph{Local lookahead and speculative decoding.}
Local future tokens have also been used to accelerate autoregressive decoding. Speculative decoding
and speculative sampling draft multiple tokens with a smaller model and verify them with a target
model, while tree-based speculative inference and lookahead decoding exploit local token trees or
parallel n-gram guesses to reduce latency
\citep{leviathan2023fast, chen2023accelerating, miao2023specinfer, fu2024break}. Recent
reasoning-oriented variants use reward models, draft reasoning chains, or semantic verification to
accelerate long chain-of-thought generation
\citep{liao2025reward, wang2025efficient, fu2025scaling}. These methods show that local futures
can be generated and checked efficiently, but mainly treat lookahead as acceleration. LBR uses
lookahead as decision evidence: sampled branches are forwarded by the same language model,
compared by a router, and pruned according to a trainable routing distribution.

\paragraph{Reinforcement learning with tree-structured exploration.}
RL with verifiable rewards has become a standard recipe for mathematical reasoning, with GRPO and
DeepSeek-R1 showing that reward-weighted sampled trajectories can elicit strong reasoning behavior
\citep{shao2024deepseekmath, guo2025deepseek}. A related line introduces tree structure into RL
training: MCTS-DPO collects step-level preference pairs with Monte Carlo tree search and updates
the policy with DPO \citep{xie2024monte}; TreeRL uses entropy-guided tree search to fork from
uncertain intermediate tokens and derive process-level supervision \citep{hou2025treerl};
DeepSearch embeds MCTS into RLVR training for exploration and credit assignment
\citep{wu2025deepsearch}; and TreePO reformulates rollouts as segment-level tree sampling with
tree-aware advantages \citep{li2025treepo}. These methods use tree search primarily to produce
better training traces, preferences, or advantage estimates over coarse reasoning units. LBR differs in
that the tree is itself the learned test-time decoding policy: the same local tree serves as the
inference-time scaling mechanism and the object optimized by verifier-based RL.

\paragraph{Continuous and soft-token branch-and-merge reasoning.}
Another close line of work seeks token-level width by replacing a discrete next token with a
continuous or soft representation. Soft Thinking forms probability-weighted mixtures of token
embeddings, and follow-up analyses study when such mixtures behave like genuine parallel reasoning
versus a dominant single path \citep{zhang2025soft, wu2025llms}. Soft-token RL methods add
stochasticity and train continuous chain-of-thought processes with reinforcement learning
\citep{butt2025soft}. Multiplex Thinking is especially close: it samples multiple candidate tokens and
merges their embeddings into one continuous token with a tractable rollout probability
\citep{tang2026multiplex}. These methods occupy the same token-level branch-and-merge regime as
LBR, but differ in representation: soft-token methods compress alternatives into a shared continuous
state, whereas LBR preserves candidates as discrete forwarded branches and routes among their
hidden states.

\paragraph{Interpreting reasoning representations.}
Hao et al. \citep{hao2024training} empirically observe that continuous thoughts can retain multiple
candidate next steps and exhibit a breadth-first-search-like reasoning pattern. Building on this observation, subsequent analyses
explain the advantage of continuous reasoning primarily through constructive
expressivity results. Zhu et al. \citep{zhu2025reasoning} construct a two-layer transformer whose
continuous thoughts superpose multiple search frontiers to solve directed
graph reachability, while Gozeten et al. \citep{gozeten2025continuous} construct a one-layer continuous-CoT
transformer that aggregates multiple partial trajectories to solve a
subset-sum-type combinatorial problem. These works highlight how a
shared continuous representation can support parallel exploration. We study a
complementary question: whether different reasoning frameworks preserve
task-relevant branch information clearly enough for downstream decisions.
Linear probing fits a lightweight classifier on frozen hidden states to test
whether a target property is linearly decodable
\citep{alain2016understanding,hewitt-liang-2019-designing,
belinkov-2022-probing, azaria-mitchell-2023-internal, 
jin-etal-2025-exploring}. Using probes as a diagnostic, we
provide a representation-level explanation for the performance differences
among discrete CoT, soft-token reasoning, and LBR: forwarding candidate
branches makes the correct target more recoverable in LBR's post-candidate
states, whereas merging candidates into soft states blurs concept identity.

\section{Conclusion}

We presented Local Branch Routing (LBR), a trainable test-time scaling framework that maintains a
rolling local lookahead tree during autoregressive decoding. By forwarding candidate branches before
routing, LBR lets each token decision use post-candidate hidden states rather than only the root
next-token distribution. The resulting prune--shift--grow process preserves discrete branch identities,
reuses surviving local futures, and defines a tractable tree-trajectory likelihood for RLVR training.
Synthetic hierarchical-planning experiments show that these forwarded branch states provide useful
decision evidence and avoid the concept ambiguity of soft-token mixtures. Math-reasoning results
further show that LBR improves over discrete CoT, vanilla discrete-token RLVR, and soft-token
branching baselines.

The main limitation is locality. LBR uses shallow branch information, so its predictability is limited
to evidence exposed within the local lookahead window. It does not realize full long-range planning
or global search over complete solutions. Larger lookahead depth can strengthen the routing signal,
but at higher forwarding cost, creating a trade-off between speed and effect. Future work may combine
local branch routing with adaptive depth, selective global search, or learned triggers for when longer
planning is needed.

\bibliographystyle{plainnat}
\bibliography{reference}

@article{zhang2025soft,
  title={Soft thinking: Unlocking the reasoning potential of llms in continuous concept space},
  author={Zhang, Zhen and He, Xuehai and Yan, Weixiang and Shen, Ao and Zhao, Chenyang and Wang, Shuohang and Shen, Yelong and Wang, Xin Eric},
  journal={arXiv preprint arXiv:2505.15778},
  year={2025}
}

@article{gozeten2025continuous,
  title={Continuous chain of thought enables parallel exploration and reasoning},
  author={Gozeten, Halil Alperen and Ildiz, M Emrullah and Zhang, Xuechen and Harutyunyan, Hrayr and Rawat, Ankit Singh and Oymak, Samet},
  journal={arXiv preprint arXiv:2505.23648},
  year={2025}
}

@inproceedings{jin-etal-2025-exploring,
    title = "Exploring Concept Depth: How Large Language Models Acquire Knowledge and Concept at Different Layers?",
    author = "Jin, Mingyu  and
      Yu, Qinkai  and
      Huang, Jingyuan  and
      Zeng, Qingcheng  and
      Wang, Zhenting  and
      Hua, Wenyue  and
      Zhao, Haiyan  and
      Mei, Kai  and
      Meng, Yanda  and
      Ding, Kaize  and
      Yang, Fan  and
      Du, Mengnan  and
      Zhang, Yongfeng",
    editor = "Rambow, Owen  and
      Wanner, Leo  and
      Apidianaki, Marianna  and
      Al-Khalifa, Hend  and
      Eugenio, Barbara Di  and
      Schockaert, Steven",
    booktitle = "Proceedings of the 31st International Conference on Computational Linguistics",
    month = jan,
    year = "2025",
    address = "Abu Dhabi, UAE",
    publisher = "Association for Computational Linguistics",
    url = "https://aclanthology.org/2025.coling-main.37/",
    pages = "558--573"
}

@inproceedings{azaria-mitchell-2023-internal,
    title = "The Internal State of an {LLM} Knows When It{'}s Lying",
    author = "Azaria, Amos  and
      Mitchell, Tom",
    editor = "Bouamor, Houda  and
      Pino, Juan  and
      Bali, Kalika",
    booktitle = "Findings of the Association for Computational Linguistics: EMNLP 2023",
    month = dec,
    year = "2023",
    address = "Singapore",
    publisher = "Association for Computational Linguistics",
    url = "https://aclanthology.org/2023.findings-emnlp.68/",
    doi = "10.18653/v1/2023.findings-emnlp.68",
    pages = "967--976"
}

@article{belinkov-2022-probing,
    title = "Probing Classifiers: Promises, Shortcomings, and Advances",
    author = "Belinkov, Yonatan",
    journal = "Computational Linguistics",
    volume = "48",
    number = "1",
    month = mar,
    year = "2022",
    address = "Cambridge, MA",
    publisher = "MIT Press",
    url = "https://aclanthology.org/2022.cl-1.7/",
    doi = "10.1162/coli_a_00422",
    pages = "207--219"
}

@inproceedings{hewitt-liang-2019-designing,
    title = "Designing and Interpreting Probes with Control Tasks",
    author = "Hewitt, John  and
      Liang, Percy",
    editor = "Inui, Kentaro  and
      Jiang, Jing  and
      Ng, Vincent  and
      Wan, Xiaojun",
    booktitle = "Proceedings of the 2019 Conference on Empirical Methods in Natural Language Processing and the 9th International Joint Conference on Natural Language Processing (EMNLP-IJCNLP)",
    month = nov,
    year = "2019",
    address = "Hong Kong, China",
    publisher = "Association for Computational Linguistics",
    url = "https://aclanthology.org/D19-1275/",
    doi = "10.18653/v1/D19-1275",
    pages = "2733--2743"
}

@article{alain2016understanding,
  title={Understanding intermediate layers using linear classifier probes},
  author={Alain, Guillaume and Bengio, Yoshua},
  journal={arXiv preprint arXiv:1610.01644},
  year={2016}
}

@article{wu2025llms,
  title={Llms are single-threaded reasoners: Demystifying the working mechanism of soft thinking},
  author={Wu, Junhong and Lu, Jinliang and Ren, Zixuan and Hu, Gangqiang and Wu, Zhi and Dai, Dai and Wu, Hua},
  journal={arXiv preprint arXiv:2508.03440},
  year={2025}
}

@article{butt2025soft,
  title={Soft tokens, hard truths},
  author={Butt, Natasha and Kwiatkowski, Ariel and Labiad, Ismail and Kempe, Julia and Ollivier, Yann},
  journal={arXiv preprint arXiv:2509.19170},
  year={2025}
}

@article{tang2026multiplex,
  title={Multiplex Thinking: Reasoning via Token-wise Branch-and-Merge},
  author={Tang, Yao and Dong, Li and Hao, Yaru and Dong, Qingxiu and Wei, Furu and Gu, Jiatao},
  journal={arXiv preprint arXiv:2601.08808},
  year={2026}
}

@misc{deepscaler2025,
  title={DeepScaleR: Surpassing O1-Preview with a 1.5B Model by Scaling RL},
  author={Michael Luo and Sijun Tan and Justin Wong and Xiaoxiang Shi and William Y. Tang and Manan Roongta and Colin Cai and Jeffrey Luo and Li Erran Li and Raluca Ada Popa and Ion Stoica},
  howpublished={\url{https://pretty-radio-b75.notion.site/DeepScaleR-Surpassing-O1-Preview-with-a-1-5B-Model-by-Scaling-RL-19681902c1468005bed8ca303013a4e2}},
  note={Notion Blog},
  year={2025}
}

@article{hao2024training,
  title={Training large language models to reason in a continuous latent space},
  author={Hao, Shibo and Sukhbaatar, Sainbayar and Su, DiJia and Li, Xian and Hu, Zhiting and Weston, Jason and Tian, Yuandong},
  journal={arXiv preprint arXiv:2412.06769},
  year={2024}
}

@article{wei2022chain,
  title={Chain-of-thought prompting elicits reasoning in large language models},
  author={Wei, Jason and Wang, Xuezhi and Schuurmans, Dale and Bosma, Maarten and Xia, Fei and Chi, Ed and Le, Quoc V and Zhou, Denny and others},
  journal={Advances in neural information processing systems},
  volume={35},
  pages={24824--24837},
  year={2022}
}

@article{snell2024scaling,
  title={Scaling llm test-time compute optimally can be more effective than scaling model parameters},
  author={Snell, Charlie and Lee, Jaehoon and Xu, Kelvin and Kumar, Aviral},
  journal={arXiv preprint arXiv:2408.03314},
  year={2024}
}

@article{wang2022self,
  title={Self-consistency improves chain of thought reasoning in language models},
  author={Wang, Xuezhi and Wei, Jason and Schuurmans, Dale and Le, Quoc and Chi, Ed and Narang, Sharan and Chowdhery, Aakanksha and Zhou, Denny},
  journal={arXiv preprint arXiv:2203.11171},
  year={2022}
}

@inproceedings{hao2023reasoning,
  title={Reasoning with language model is planning with world model},
  author={Hao, Shibo and Gu, Yi and Ma, Haodi and Hong, Joshua and Wang, Zhen and Wang, Daisy and Hu, Zhiting},
  booktitle={Proceedings of the 2023 Conference on Empirical Methods in Natural Language Processing},
  pages={8154--8173},
  year={2023}
}

@article{wang2025efficient,
  title={Efficient reasoning for llms through speculative chain-of-thought},
  author={Wang, Jikai and Li, Juntao and Hou, Jianye and Yan, Bowen and Wu, Lijun and Zhang, Min},
  journal={arXiv preprint arXiv:2504.19095},
  year={2025}
}

@article{fu2024break,
  title={Break the sequential dependency of llm inference using lookahead decoding},
  author={Fu, Yichao and Bailis, Peter and Stoica, Ion and Zhang, Hao},
  journal={arXiv preprint arXiv:2402.02057},
  year={2024}
}

@article{miao2023specinfer,
  title={Specinfer: Accelerating generative large language model serving with tree-based speculative inference and verification},
  author={Miao, Xupeng and Oliaro, Gabriele and Zhang, Zhihao and Cheng, Xinhao and Wang, Zeyu and Zhang, Zhengxin and Wong, Rae Ying Yee and Zhu, Alan and Yang, Lijie and Shi, Xiaoxiang and others},
  journal={arXiv preprint arXiv:2305.09781},
  year={2023}
}

@article{chen2023accelerating,
  title={Accelerating large language model decoding with speculative sampling},
  author={Chen, Charlie and Borgeaud, Sebastian and Irving, Geoffrey and Lespiau, Jean-Baptiste and Sifre, Laurent and Jumper, John},
  journal={arXiv preprint arXiv:2302.01318},
  year={2023}
}

@article{xie2024monte,
  title={Monte carlo tree search boosts reasoning via iterative preference learning},
  author={Xie, Yuxi and Goyal, Anirudh and Zheng, Wenyue and Kan, Min-Yen and Lillicrap, Timothy P and Kawaguchi, Kenji and Shieh, Michael},
  journal={arXiv preprint arXiv:2405.00451},
  year={2024}
}

@article{guo2025deepseek,
  title={Deepseek-r1: Incentivizing reasoning capability in llms via reinforcement learning},
  author={Guo, Daya and Yang, Dejian and Zhang, Haowei and Song, Junxiao and Wang, Peiyi and Zhu, Qihao and Xu, Runxin and Zhang, Ruoyu and Ma, Shirong and Bi, Xiao and others},
  journal={arXiv preprint arXiv:2501.12948},
  year={2025}
}

@article{fu2025scaling,
  title={Scaling speculative decoding with lookahead reasoning},
  author={Fu, Yichao and Ge, Rui and Shao, Zelei and Deng, Zhijie and Zhang, Hao},
  journal={arXiv preprint arXiv:2506.19830},
  year={2025}
}

@article{liao2025reward,
  title={Reward-guided speculative decoding for efficient llm reasoning},
  author={Liao, Baohao and Xu, Yuhui and Dong, Hanze and Li, Junnan and Monz, Christof and Savarese, Silvio and Sahoo, Doyen and Xiong, Caiming},
  journal={arXiv preprint arXiv:2501.19324},
  year={2025}
}

@inproceedings{leviathan2023fast,
  title={Fast inference from transformers via speculative decoding},
  author={Leviathan, Yaniv and Kalman, Matan and Matias, Yossi},
  booktitle={International Conference on Machine Learning},
  pages={19274--19286},
  year={2023},
  organization={PMLR}
}

@article{yao2023tree,
  title={Tree of thoughts: Deliberate problem solving with large language models},
  author={Yao, Shunyu and Yu, Dian and Zhao, Jeffrey and Shafran, Izhak and Griffiths, Tom and Cao, Yuan and Narasimhan, Karthik},
  journal={Advances in neural information processing systems},
  volume={36},
  pages={11809--11822},
  year={2023}
}

@article{inoue2025wider,
  title={Wider or deeper? scaling llm inference-time compute with adaptive branching tree search},
  author={Inoue, Yuichi and Misaki, Kou and Imajuku, Yuki and Kuroki, So and Nakamura, Taishi and Akiba, Takuya},
  journal={arXiv preprint arXiv:2503.04412},
  year={2025}
}

@inproceedings{muennighoff2025s1,
  title={s1: Simple test-time scaling},
  author={Muennighoff, Niklas and Yang, Zitong and Shi, Weijia and Li, Xiang Lisa and Fei-Fei, Li and Hajishirzi, Hannaneh and Zettlemoyer, Luke and Liang, Percy and Cand{\`e}s, Emmanuel and Hashimoto, Tatsunori B},
  booktitle={Proceedings of the 2025 Conference on Empirical Methods in Natural Language Processing},
  pages={20286--20332},
  year={2025}
}

@article{li2025policy,
  title={Policy guided tree search for enhanced llm reasoning},
  author={Li, Yang},
  journal={arXiv preprint arXiv:2502.06813},
  year={2025}
}

@article{zhou2023language,
  title={Language agent tree search unifies reasoning acting and planning in language models},
  author={Zhou, Andy and Yan, Kai and Shlapentokh-Rothman, Michal and Wang, Haohan and Wang, Yu-Xiong},
  journal={arXiv preprint arXiv:2310.04406},
  year={2023}
}

@article{wu2025deepsearch,
  title={DeepSearch: Overcome the Bottleneck of Reinforcement Learning with Verifiable Rewards via Monte Carlo Tree Search},
  author={Wu, Fang and Xuan, Weihao and Qi, Heli and Lu, Ximing and Tu, Aaron and Li, Li Erran and Choi, Yejin},
  journal={arXiv preprint arXiv:2509.25454},
  year={2025}
}

@article{li2025treepo,
  title={Treepo: Bridging the gap of policy optimization and efficacy and inference efficiency with heuristic tree-based modeling},
  author={Li, Yizhi and Gu, Qingshui and Wen, Zhoufutu and Li, Ziniu and Xing, Tianshun and Guo, Shuyue and Zheng, Tianyu and Zhou, Xin and Qu, Xingwei and Zhou, Wangchunshu and others},
  journal={arXiv preprint arXiv:2508.17445},
  year={2025}
}

@inproceedings{hou2025treerl,
  title={Treerl: Llm reinforcement learning with on-policy tree search},
  author={Hou, Zhenyu and Hu, Ziniu and Li, Yujiang and Lu, Rui and Tang, Jie and Dong, Yuxiao},
  booktitle={Proceedings of the 63rd Annual Meeting of the Association for Computational Linguistics (Volume 1: Long Papers)},
  pages={12355--12369},
  year={2025}
}

@article{zhu2025reasoning,
  title={Reasoning by superposition: A theoretical perspective on chain of continuous thought},
  author={Zhu, Hanlin and Hao, Shibo and Hu, Zhiting and Jiao, Jiantao and Russell, Stuart and Tian, Yuandong},
  journal={arXiv preprint arXiv:2505.12514},
  year={2025}
}

@article{shao2024deepseekmath,
  title={Deepseekmath: Pushing the limits of mathematical reasoning in open language models},
  author={Shao, Zhihong and Wang, Peiyi and Zhu, Qihao and Xu, Runxin and Song, Junxiao and Bi, Xiao and Zhang, Haowei and Zhang, Mingchuan and Li, YK and Wu, Yang and others},
  journal={arXiv preprint arXiv:2402.03300},
  year={2024}
}

\newpage
\appendix

\section{Complete Decoding Framework}
\subsection{Rolling Local Lookahead Tree}

Let $x_{<t}$ denote the committed prefix before decoding position $t$. Standard autoregressive
decoding samples the next token directly from $\pi_\theta(\cdot \mid x_{<t})$. In contrast, LBR first
constructs a local lookahead tree rooted at $x_{<t}$.

Fix a tree depth $L \geq 1$ and width $K \geq 1$. At decoding step $t$, the local lookahead tree is
denoted by
\[
    \mathcal{A}_t = (\mathcal{V}_t,\mathcal{E}_t,r_t),
\]
where $r_t$ is the root corresponding to the committed prefix $x_{<t}$, $\mathcal{V}_t$ is the set of
tree nodes, and $\mathcal{E}_t$ is the set of directed parent--child edges. Every non-leaf node has
$K$ sampled children, and every root-to-leaf path has length $L$. Thus, $\mathcal{A}_t$ is a
depth-$L$, width-$K$ sampled tree.

Each node $v\in\mathcal{V}_t\setminus\{r_t\}$ carries a token label
\[
    \mathrm{tok}(v)\in\mathcal{V},
\]
a depth $d(v)\in\{1,\ldots,L\}$, and a path from the root
\[
    \mathrm{path}(v)
    =
    \bigl(
    \mathrm{tok}(v_1),\ldots,\mathrm{tok}(v_{d(v)})
    \bigr),
\]
where $r_t \to v_1 \to \cdots \to v_{d(v)}=v$ is the unique path from the root to $v$.
Children are sampled autoregressively: for every node $v$ with $d(v)<L$, its children
$\mathrm{Ch}(v)=\{c_1(v),\ldots,c_K(v)\}$ are drawn from
\[
    \mathrm{tok}(c_i(v))
    \sim
    \tilde{\pi}_\theta
    \left(
        \cdot
        \mid
        x_{<t}, \mathrm{path}(v)
    \right),
    \qquad i=1,\ldots,K.
\]
Here $\tilde{\pi}_\theta$ denotes the model distribution after the same sampling filters used in
ordinary decoding, such as temperature, top-$p$, top-$k$, or other filters.
The ``Grow local tree'' panels of Figure~\ref{fig:lbr-decoding} show the resulting tree before
routing. For $L=1$, the tree contains only candidate next tokens; for $L=2$, each candidate next
token also has one layer of sampled continuations.

The crucial point is that the tree is not only sampled, but also forwarded. For every non-root node
$v\in\mathcal{V}_t$, we compute and store the hidden state
\[
    h_t(v)
    =
    f_\theta
    \left(
        x_{<t}, \mathrm{path}(v)
    \right)
    \in \mathbb{R}^D.
\]
Therefore, the local tree provides more than candidate token identities: it provides hidden-state
representations of local future continuations. These hidden states are the evidence used by the
router.

\subsection{Routing over Forwarded Subtrees}

The depth-$1$ children of the root are the roots of the candidate subtrees:
\[
    u_{t,1},\ldots,u_{t,K}
    \in
    \mathrm{Ch}(r_t).
\]
The token label $\mathrm{tok}(u_{t,k})$ is the $k$-th candidate next token. The subtree rooted at
$u_{t,k}$ is
\[
    \mathcal{A}_t[u_{t,k}]
    =
    \{v\in\mathcal{V}_t : u_{t,k} \text{ lies on the path from } r_t \text{ to } v\}.
\]
This subtree contains the candidate token $\mathrm{tok}(u_{t,k})$ and all already-forwarded local
continuations below it up to depth $L$.

Let $\rho_\phi$ be a router parameterized by $\phi$. Given the committed prefix and the forwarded
lookahead tree, the router outputs a distribution over the $K$ depth-$1$ subtrees:
\[
    \rho_\phi
    \left(
        \cdot
        \mid
        x_{<t}, \mathcal{A}_t; \theta
    \right)
    \in
    \Delta^K.
\]
We write the dependence on $\theta$ because the router input includes hidden states
$\{h_t(v):v\in\mathcal{V}_t\}$ computed by the base language model. Equivalently, one may denote
this dependence by $\rho_{\phi,\theta}(\cdot\mid x_{<t},\mathcal{A}_t)$.

The router samples or selects a subtree index
\[
    k_t^\star
    \sim
    \rho_\phi
    \left(
        \cdot
        \mid
        x_{<t}, \mathcal{A}_t; \theta
    \right),
\]
and commits the corresponding root token
\[
    x_t
    =
    \mathrm{tok}(u_{t,k_t^\star}).
\]
The ``Router selects subtree'' panels in Figure~\ref{fig:lbr-decoding} emphasize that the router
scores candidate subtrees using the forwarded hidden states of the local tree, rather than using only
the root next-token logits.
\subsection{Efficient Tree-Structured Forwarding}

A naive implementation would forward every sampled path independently. LBR instead uses a tree representation that shares common prefixes among sampled paths. The sampled paths are stored both as a full tree with multiplicity and as a deduplicated trie. The full tree preserves the sampling multiplicity of repeated tokens, while the trie allows shared prefixes to reuse key-value cache entries.

At each grow step, all current frontier nodes are processed in a batched tree-attention forward pass. Each frontier node attends to the committed prefix and to the ancestor tokens along its own branch:
\[
    \mathrm{KV}(v)
    =
    \mathrm{KV}(x_{<t})
    \;\Vert\;
    \mathrm{KV}(\mathrm{ancestors}(v)).
\]
This tree-causal attention pattern computes the hidden states for all newly expanded nodes in parallel. Since the router consumes these hidden states, no additional language-model forward pass is needed solely for routing.

The computational overhead of LBR comes from forwarding the local tree frontier and applying the lightweight router. The benefit is that each token decision is informed by a structured set of local future hidden states rather than by the root next-token distribution alone.

\subsection{Set-Attention Router}
\label{subsec: router structure}

The router compares the $K$ candidate subtrees using hidden states from the already-forwarded
lookahead tree. Since the order of sampled rollouts is arbitrary, the router should not depend on
the incidental ordering of sampled branches. We instantiate $\rho_\phi$ as a hierarchical
set-attention router with two levels: it first aggregates rollouts within each candidate subtree, and
then compares the resulting candidate representations across subtrees.

For each candidate subtree rooted at $u_{t,k}$, let
\[
    \mathcal P_{t,k}
    =
    \{p_{t,k,1},\ldots,p_{t,k,M_k}\}
\]
denote the set of root-to-leaf rollout paths contained in that subtree, where
$M_k=K^{L-1}$ for a full width-$K$, depth-$L$ tree. Each rollout path
$p_{t,k,m}$ has an ordered hidden-state sequence
\[
    H_{t,k,m}
    =
    \bigl(
    h_{t,k,m}^{(1)},\ldots,h_{t,k,m}^{(L)}
    \bigr)
    \in \mathbb R^{L\times D},
\]
where $h_{t,k,m}^{(\ell)}$ is the hidden state at depth $\ell$ along that rollout. The first hidden
state $h_{t,k,m}^{(1)}$ corresponds to the candidate root $u_{t,k}$ and is therefore shared across
all rollouts in the same subtree.

We first encode each rollout path with a small position-aware Transformer encoder. Let
$W_{\mathrm{in}}\in\mathbb R^{p\times D}$ be an input projection and let
$e_1,\ldots,e_L\in\mathbb R^p$ be learned depth embeddings. We define
\[
    \bar h_{t,k,m}^{(\ell)}
    =
    W_{\mathrm{in}}h_{t,k,m}^{(\ell)} + e_\ell,
    \qquad \ell=1,\ldots,L.
\]
After prepending a learned $\texttt{[CLS]}$ vector, we apply a shared path encoder:
\[
    R_{t,k,m}
    =
    \mathrm{PathEnc}_\phi
    \left(
        [\texttt{CLS}],
        \bar h_{t,k,m}^{(1)},\ldots,\bar h_{t,k,m}^{(L)}
    \right),
\]
and read out the $\texttt{[CLS]}$ representation as the rollout feature
\[
    r_{t,k,m}
    =
    R_{t,k,m}^{\texttt{[CLS]}}
    \in \mathbb R^p .
\]
This step preserves the ordered structure along a rollout path.

Next, we aggregate the rollout features within each candidate subtree. Since the rollouts under
the same candidate token have no canonical order, we use a set-attention block followed by
attention pooling:
\[
    Z_{t,k}
    =
    \mathrm{SetAttn}^{\mathrm{sub}}_\phi
    \left(
        r_{t,k,1},\ldots,r_{t,k,M_k}
    \right),
\]
\[
    g_{t,k}
    =
    \mathrm{AttnPool}_\phi
    \left(
        q_{\mathrm{sub}}, Z_{t,k}
    \right)
    \in \mathbb R^p ,
\]
where $q_{\mathrm{sub}}\in\mathbb R^p$ is a learned pooling query. This gives one permutation-invariant
representation $g_{t,k}$ for the entire subtree rooted at $u_{t,k}$.

Finally, the router compares the $K$ candidate subtrees jointly. We pass the candidate
representations through a second set-attention block:
\[
    \tilde g_{t,1},\ldots,\tilde g_{t,K}
    =
    \mathrm{SetAttn}^{\mathrm{cand}}_\phi
    \left(
        g_{t,1},\ldots,g_{t,K}
    \right),
\]
which is permutation-equivariant across candidate subtrees. Each candidate is then assigned a
scalar score
\[
    s_{t,k}
    =
    w_\phi^\top \tilde g_{t,k},
\]
and the router distribution is
\[
    \rho_\phi(k\mid x_{<t},\mathcal A_t;\theta)
    =
    \frac{\exp(s_{t,k}/\tau)}
    {\sum_{k'=1}^{K}\exp(s_{t,k'}/\tau)} ,
\]
where $\tau>0$ is a routing temperature.

For the main setting $L=1$, each candidate subtree contains only a single forwarded token, so
$M_k=1$ and the path encoder receives only the hidden state $h_t(u_{t,k})$. In this case,
within-subtree aggregation is trivial, but the cross-subtree set-attention block still allows the
router to compare the post-token hidden states of the $K$ sampled candidate next tokens. For
$L>1$, the same architecture first summarizes each local rollout path and then aggregates the
multiple rollouts under each candidate before comparing candidates.

\subsubsection{Router Hyperparameters }
\label{apd: router hyper}
We instantiate the router with internal dimension $p = 128$ and $H = 4$ attention heads in every multi-head-attention block. The input
  projection $W_{\mathrm{in}} \in \mathbb{R}^{p \times D}$ uses the language model's hidden dimension $D = 1536$                              
  (DeepSeek-R1-Distill-Qwen-1.5B). The shared path encoder $\mathrm{PathEnc}_\phi$ is a single pre-LN Transformer encoder layer with GELU  
  activations and FFN expansion factor $4$. Each set-attention block $\mathrm{SetAttn}^{\mathrm{sub}}_\phi$ and
$\mathrm{SetAttn}^{\mathrm{cand}}_\phi$ is one SAB layer with the same head count and FFN structure. The within-subtree pooling
  $\mathrm{AttnPool}_\phi$ uses a single learned query $q_{\mathrm{sub}} \in \mathbb{R}^p$ via standard scaled-dot-product attention. We
  disable dropout throughout the router. The final routing softmax temperature is $\tau = 1$. The learning rate of router is $1 \times 10^{-4}$.

\section{Why Soft Thinking Underperforms in Hierarchical Planning: Concept Ambiguity}
\label{apd: soft ambig}
\begin{figure}[t]
    \centering
    \includegraphics[width=0.42\textwidth]{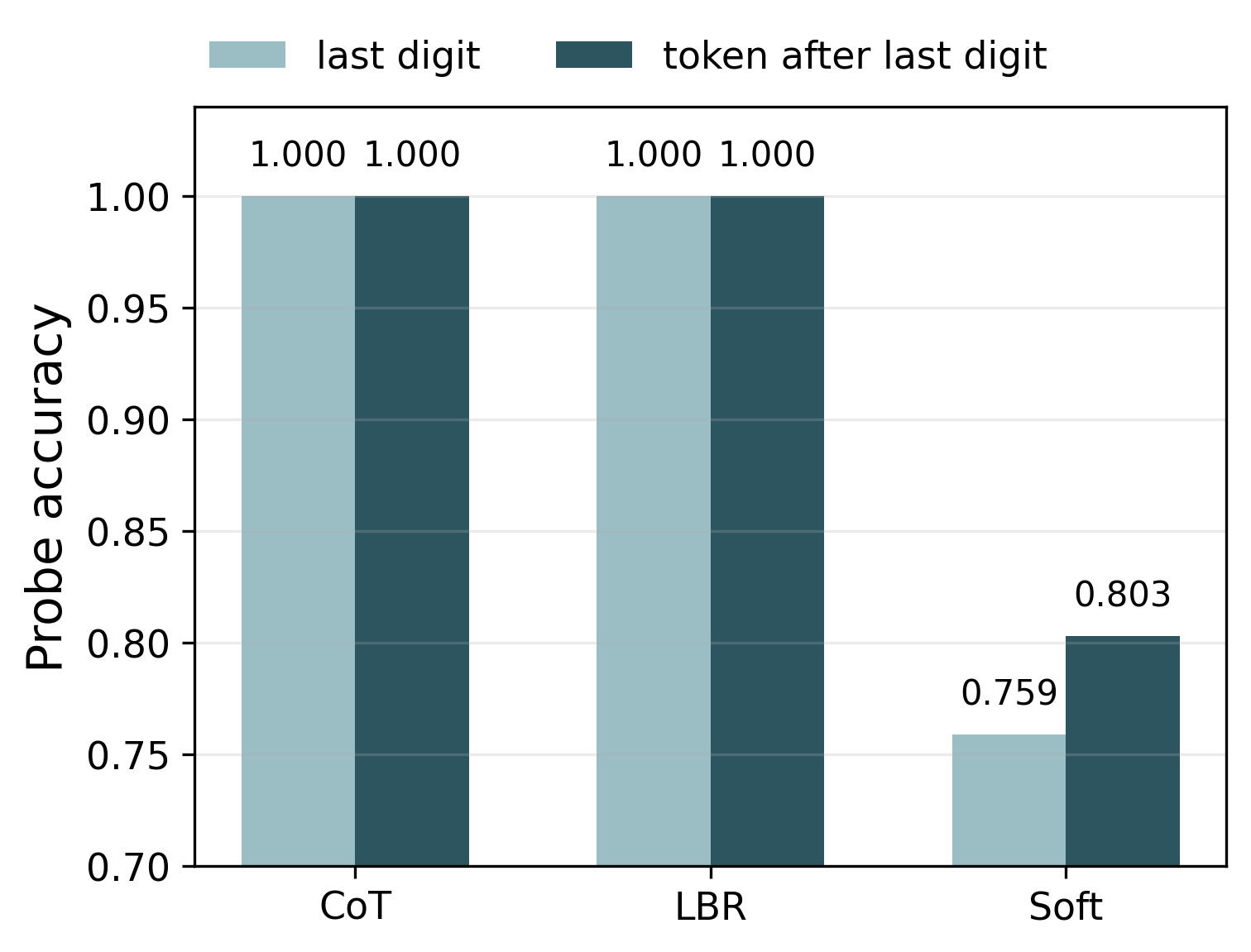}
    \caption{
    Concept-identity probe after the first generated graph node. Discrete CoT and LBR
preserve node identity through discrete one-hot digit tokens, while Soft Thinking loses concept
information after mixing candidate embeddings.
    }
    \label{fig:n1-identity-probe}
\end{figure}

Mixture embeddings can blur the identity of the concept branch being followed. In our task, discrete CoT and LBR always condition on
one-hot digit tokens. Soft Thinking instead feeds mixtures of legal candidate embeddings at branching positions,
which can leave the model in an ambiguous state between multiple possible concepts.

Figure~\ref{fig:n1-identity-probe} validates this effect with a concept-identity probe. After the model
processes the digit sequence of the first generated graph node, a linear probe can perfectly recover
the node identity for Discrete CoT and LBR. For Soft Thinking, however, the probe remains
substantially below perfect accuracy, reaching only \(0.759\) at the final digit of the node and
\(0.803\) at the following arrow token. Thus, even after a full node has been processed, the soft
hidden state does not fully identify which concept branch the model is following.

This supports the mechanism suggested in Figure~\ref{fig:interp}: LBR preserves
candidate-specific branches and routes among them, whereas Soft Thinking compresses competing
candidates into a shared continuous state. The resulting concept ambiguity explains why Soft
Thinking underperforms LBR on the radix-translated planning task.
\end{document}